# Unsupervised ensemble of experts (EoE) framework for automatic binarization of document images

Reza Farrahi Moghaddam, Fereydoun Farrahi Moghaddam and Mohamed Cheriet
Synchromedia Laboratory, École de technologie supérieure, Montreal (QC), Canada H3C 1K3
Email: imriss@ieee.org, rfarrahi@synchromedia.ca, ffarrahi@synchromedia.ca, mohamed.cheriet@etsmtl.ca
Tel.: +1(514)396-8972, Fax: +1(514)396-8595

*Abstract*—In recent years, a large number of binarization methods have been developed, with varying performance generalization and strength against different benchmarks. In this work, to leverage on these methods, an ensemble of experts (EoE) framework is introduced, to efficiently combine the outputs of various methods. The proposed framework offers a new selection process of the binarization methods, which are actually the experts in the ensemble, by introducing three concepts: confidentness, endorsement and schools of experts. The framework, which is highly objective, is built based on two general principles: (i) consolidation of saturated opinions and (ii) identification of schools of experts. After building the endorsement graph of the ensemble for an input document image based on the confidentness of the experts, the saturated opinions are consolidated, and then the schools of experts are identified by thresholding the consolidated endorsement graph. A variation of the framework, in which no selection is made, is also introduced that combines the outputs of all experts using endorsement-dependent weights. The EoE framework is evaluated on the set of participating methods in the H-DIBCO'12 contest and also on an ensemble generated from various instances of grid-based Sauvola method with promising performance.

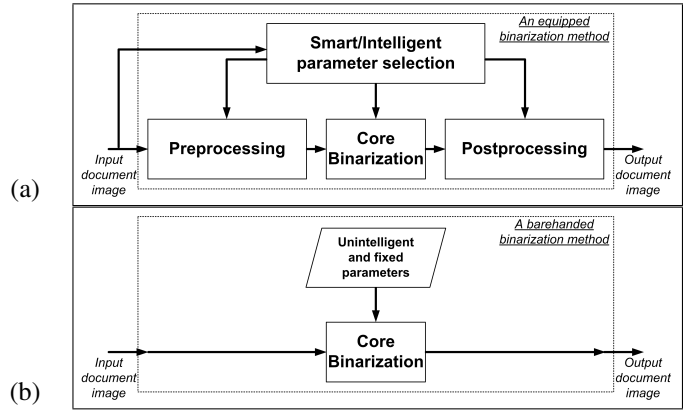

Fig. 1. a) A schematic anatomy of a binarization method. b) A barehanded binarization method.

## I. INTRODUCTION

Binarization of document images is a critical step in many document processing work-flows, and at the same time it is a good example of a complex analysis problem [1], [2]. In a highly-simplified representation, it can be represented by a process consisting of four subprocesses (as shown in Figure 1(a)). The core binarization subprocess (for example, grid-based Sauvola formula, equation (7) in [3]) constitutes only a fraction of the whole binarization method. The three other subprocesses are preprocessing [4] and postprocessing [5], as well smart and intelligent selection of the parameters, where the latter is key to success and robustness of the overall method [6]. It would be an unfair evaluation if a barehanded method as shown in Figure 1(b), for example Sauvola's formula alone, is compared with an equipped method with preprocessing and postprocessing. We hope this help to raise the concern about high-degree of ambiguity in evaluation protocols used in publications. This has been resulted in using some well-known methods being used in their barehanded forms as inferior examples in the comparisons. In turn, this has resulted in an endless search for more complicated methods, where this complexity usually achieved in pre- and post-processing steps, which could be easily combined with those dislike methods to achieve higher performance even compared to those of the newer methods. For example, we have shown in a recent paper that a classic method with automatic parameter optimization could outperform state of the art methods. The question of objectivity and fairness in comparison of methods is beyond the scope of this work, and we hope the community pursues discussions toward objective, solid and practical evaluation frameworks.

Similar to many other image processing problems, the key obstacle in front of binarization methods is the 2-dimensional (2D) nature of images. While these objects are highly rich in terms of spatial relations, there is no standard way to convert them into 1-dimensional (1D) feature vectors. This problem roots in the lack of a proper "order" in $\mathbb{R}^2$. This has led to a huge set of binarization methods which try any possible approach from Laplacian energy [7] and super resolution [8] to digging into image data and building text extraction models [1], [4], [5], [9], [10].

While development of new methods will continue, combination of already proven methods has been also considered [4], [9], [10]. There are two main trends along this direction: 1) to combine the outputs of various methods [9] and 2) to use the outputs of some of them in preprocessing/postprocessing of the others [4], [10]. Although this has been successful for small number of methods, its advantage cannot be guaranteed when the number of methods grows; if the number of similar methods become higher than the others, they could introduce bias in the overall decision. This is our main motivation in this work to develop a framework in which a dependable combination of experts can be achieved with minimal subjective intervention.

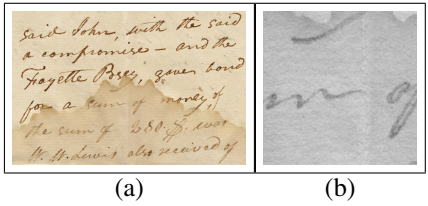

Fig. 2. An example from H-DIBCO'12 dataset [11]. a) The original image. b) An extract from (a) used in the examples.

In this work, an ensemble-of-experts (EoE) framework is proposed which tries to identify the right experts from an ensemble of experts for each input image using the concept of endorsement. First, the confidentness map of a binarization methods is defined. Then, the endorsement concept is introduced for an expert. Next, the endorsement graph among the experts is defined. Based on a few general principles considered to be applicable to ensembles of experts, the endorsement graph is analyzed in order to identify the right experts.

It is worth noting that experts are different from classifiers. First, at each time, an expert works on a set of similar problems, and uses its collected information to adjust its parameters for that set of problems. In this way, an expert is more similar to a meta-classifier. Second, and more importantly, in contrast to classifiers, which start from the feature vectors extracted from the objects, experts work directly on the object, and feature extraction (if present) is actually a part of their analysis. Although many experts could be decomposed into a feature extractor and a classifier, we consider an expert as a black box. In this work, we choose an ensemble of experts framework over an ensemble of classifiers framework because it allows more independence and diversity among the experts.[1]

The proposed framework along with a parametric binarization method can be implicitly seen as a featureless binarization approach to document images. The robustness of the EoE framework is especially important when the ensemble of the experts have a large number of common experts. In addition, the proposed EoE framework has the capacity to be applied to other decision-making problems.

The paper is organized as follows. The proposed EoE framework is described in Section II by introducing the confidentness map, endorsement graph, and school of experts. The application of the proposed framework on the H-DIBCO'12 methods is provided in Section III-A. An automatic binarization method based on the grid-based Sauvola method and the EoE framework is presented in Section III-B, and is evaluated on the H-DIBCO'12 dataset. The conclusion and future prospects are discussed in Section IV.

---

[1]Also, note that the weak/strong attributes do not apply to experts because an expert could perform well on a set of problems, while it could be a not-a-good expert for another set.

## II. THE PROPOSED ENSEMBLE-OF-EXPERTS (EoE) FRAMEWORK

Before describing the framework, it is worth noting that the process provided below is repeated for each input document image. In other words, the right experts for one image may be not proper for another image. The notation is presented below in a general manner in order to facilitate application of this framework to other decision making problems.

Assume a set of highly-correlated binary problems $\{p_i\}_{i=1}^{N_p}$. An expert $E_\omega$ is supposed to provide a binary decision $d_{i;\omega} \in \{0,1\}$ for each problem $p_i$. The ensemble of experts is denoted $\mathbf{E} = \{E_\omega\}_{\omega=1}^{N_e}$. Each expert also provides a confidentness value between zero and one for each of its decisions, denoted $c_{i;\omega}$ for the expert $E_\omega$ on the decision $d_{i;\omega}$.

If the domain of problems $i$ is an image domain $\Omega$, we call $D_\omega = \{d_{i;\omega}\} = \{d_{(k,l);\omega}\}_{(k,l)}$ the associated decision map of the expert $E_\omega$ on $\Omega$, where $i = (k,l)$ is a pixel on $\Omega$. In the same way, $C_\omega = \{c_{i;\omega}\} = \{c_{(k,l);\omega}\}_{(k,l)}$ is called the associated confidentness map of $E_\omega$. The goal is to identify the set of right experts, $\mathcal{E}$, as a subset of $E$ for each set of binary problems (or the input image in the case of binarization) that provides a better performance on that set of binary problems (input image) compared to any other possible subset selected from $E$.

In the case of binarization methods, the set of problems is the set of image pixels. Traditionally, binarization methods only provide their decision on the pixels without any estimation of the confidentness value. Therefore, before discussing furthermore the EoE framework, an approach to calculate the confidentness map of a binarization method on an image is provided in the next subsection. It is also suggested to generate the confidentness map as the secondary output of binarization methods developed in future to avoid this estimated $C$.

### A. Confidentness maps

Assuming that the binarization method $E_\omega$ provides the binary output image $D$ for the input image $I$ on the domain $\Omega$ of $n \times m$ pixels, its confidentness map for $I$ is estimated as follows. It is worth noting that $d_{i;\omega}$, which is the output of $E_\omega$ for pixel $i$, corresponds to $D_{l,k}$ pixel on the binary map where $i = (k,l)$.

In order to include the spatial relations within the confidentness map, the local estimation of stroke width on the $D$ is considered. In other words, a map on the same domain as $I$ is created that give the estimated value of the stroke width at each pixel. This map called $W$ is estimated using the grid-based modeling [3] of the stroke width algorithm [5] on a scale relative to the estimated line height. As the input $I$ could be practically a part of a document image that may leads to improper estimation of the line height using standard methods [5], the line height is estimated using a model based on the number of connected components presented on $D$:

$$G_s = \max\left(40, \left[\frac{1}{2}\sqrt{\frac{n \times m}{\min\left(400, \max\left(1, N_{cc}\right)\right)}}\right]\right), \quad (1)$$

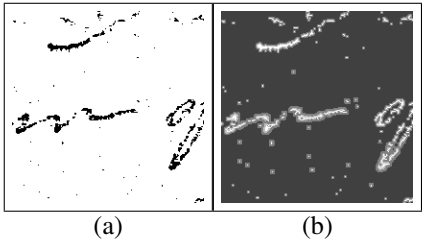

Fig. 3. An example of $C$ map. a) The $D$ map. b) The corresponding $C$ map calculated based on Algorithm 1.

where $G_s$ is the grid scale and $N_{cc}$ is the number of connected components on $D$. The parameters used in the equation are intuitively selected, and can be optimized for better performance. The complete procedure to calculate the confidentness map is provided in Algorithm 1. A mosaic grid of an image is a set of sliding patches of size defined by the grid size that have a certain degree of overlapping. Although transforming an image into its mosaic grid is straightforward, the reverse transform is not unique. Usually, average function is used to combine the values from various patches on a shared region. In Algorithm 1, we used the minimum function because of the nature of the confidentness map. Also, the steps of the confidentness map of a mosaic patch are intuitively selected to give more weight to the inner parts of strokes. An example of $C$ map is shown in Figure 3 along with its corresponding $D$.

**Algorithm 1** Confidentness map algorithm

1: **procedure** CONFIDENTNESS MAP($D$)
2:   Calculate the local stroke width map $W$ using grid-based modeling and equation (1) for $G_s$.
3:   Calculate the maximum value on $W$: $w_{G_s,max}$; correct $G_s$: $G_s := \max(4w_{G_s,max}+1, G_s)$; recalculate $W$ if necessary.
4:   Convert $D$ and $W$ to their mosaic grid equivalents with 50% overlapping: $\{D_{\text{Mg},k}\}_k$ and $\{W_{\text{Mg},k}\}_k$.
5:   For each mosaic patch $W_{\text{Mg},k}$, calcualble the maximum stroke width: $w_{G_s,\text{Mg},k}$. Using morphological operators, calculate the confidentness of that mosaic patch $C_{\text{Mg},k}$, in which pixels in the distance of $w_{G_s,\text{Mg},k}/4$ to the edges on $D_{\text{Mg},k}$ receive a value of 0.50 and 0.75 depending being outside or inside the text region. Other pixels will receive a value of 0.25 or 1.0 depending on being background or text.
6:   Combine back $\{C_{\text{Mg},k}\}_k$ using the minimum function on the overlapping regions to construct $C$. **return** $C$
7: **end procedure**

### B. Endorsement and endorsement graph

Each expert is assumed to endorse the others depending on how much their decisions on the set of problems (pixels) look similar. The endorsement of the expert $E_\alpha$ given by the expert $E_\beta$ is denoted by $R_{\alpha,\beta}$. We define $R_{\alpha,\beta}$ as

$$R_{\alpha,\beta} = \frac{\sum_i C_{i;\alpha,\text{masked},\beta}}{\sum_i C_{i;\beta}}, \quad (2)$$

where $C_{\alpha,\text{masked},\beta}$ is the masked $C_\alpha$ by $C_\beta$ defined as:

$$C_{i;\alpha,\text{masked},\beta} = \begin{cases} C_{i;\alpha} & \text{If } C_{i;\alpha} < C_{i;\beta}, \\ 0 & \text{Otherwise.} \end{cases} \quad (3)$$

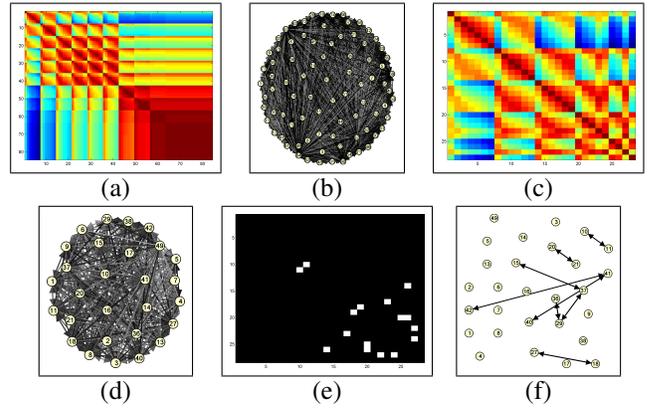

Fig. 4. An example of the endorsement graph. a) The endorsement weights $R$. b) The graph visualization of (a). c) The endorsement weights of consolidated $R$: $R'$. d) The graph visualization of (c). e) The selected schools of experts. f) The graph visualization of (e).

The $R_{\alpha,\beta}$ values can be imagined as the weights on edges of an extraverted bidirected graph. We call this graph the *endorsement graph*. An example of the endorsement graph is shown in Figure 4(b).[2] For more information on the experts used in this example see Section III-B. Also, based on $R$, a endorsement weight is assigned to each expert. The endorsement weight vector, denoted $r$, is defined as the sum of all endorsement an expert receives from the others: $r_\omega = \sum_\eta R_{\omega,\eta}$.

### C. Identification of the proper experts

The core of the EoE framework is identification of the right experts among all the ensemble's experts. In order to have a generic and generalizable approach, we set our framework on a minimal set of two common-sense principles:

1) Each set of experts which are coherent among each other form a "school" within the ensemble.
2) The less-informed experts highly endorse each other.

The first principle states that we expect a few schools of experts within the ensemble, in each one the experts considerably endorse each other. Although experts in one school agree on a large extent with each other, we still assume that their "opinions" are slightly different. This is where the second principle comes into action. Those experts that highly endorse each other most probably have a common, "saturated" opinion. Therefore, they are not proper for the image (the set of problems) under consideration.

The second principle is applied by consolidating those experts that have endorsement values higher than a threshold among themselves. In this work, we use a high threshold of $R_{\text{con;thr}} = 0.99$ for this purpose. However, this threshold value could be optimized, and more smart ways to determine it will be considered in future. From each consolidated cluster of highly-self-endorsing experts, that expert that has the highest $r$ value is kept, and the rest are removed from the ensemble. The effect of this operation on the graph of the example in

---
[2]The graphs are created using the GraphVis4Matlab toolbox, and a patch:http://www.mathworks.com/matlabcentral/fileexchange/39982-a-patch-for-graphviz4matlab-toolbox.

Figure 4 is shown in Figures 4(c) and 4(d). In the second part of expert selection, schools of experts are identified.

In this work, we assume that the experts of a school have a relatively high endorsement among each other. At the same time, in order to select the core members of a school, a variable school threshold, $R_{\text{thr}}$, is considered in such a way that the maximum size of a school will be less than or equal to 5 experts. The $R_{\text{thr}}$ is initialized using Otsu's thresholding applied to the set of endorsement values. An example of schools and selected experts is shown in Figures 4(e) and 4(f). The complete process of the EoE expert selection is provided in Algorithm 2. In each iteration in the algorithm, the threshold value $R_{\text{thr}}$ is slightly increased in order to reduce the size of schools. The geometrical increase mechanism used can be replaced by a smarter mechanism. The final selected set of experts is denoted as $\mathcal{E}$, and their number is denoted as $N_{\text{EoE}}$. The output of the EoE framework is then calculated as:

$$D_{\text{EoE}} = \frac{1}{N_{\text{EoE}}} \sum_{\omega : E_\omega \in \mathcal{E}} D_\omega. \quad (4)$$

**Algorithm 2** EoE selection algorithm

1: **procedure** EOE SELECT($R$, $R_{\text{con;thr}}$)
2:    Consolidate those experts that have endorsement higher than the threshold $R_{\text{con;thr}}$ among themselves. The new set of experts is represented by $R'$.
3:    Calculate the school threshold, $R_{\text{thr}}$, by applying Otsu's thresholding to the distribution of $R'$ values.
4:    **repeat**
5:       $R_{\text{thr}} := (1 + 2R_{\text{thr}})/3$.
6:       Calculate the binary $R'$ by applying $R_{\text{thr}}$.
7:       Identify the connected components on the binary $R'$ as current schools.
8:       Calculate the size of each school.
9:       **if** Maximum school size $< 2$ **then**
10:          Roll back to the previous $R_{\text{thr}}$ value, and exit the loop.
11:       **end if**
12:    **until** Maximum school size $<= 5$
13:    Choose experts belonging to all the schools corresponding to $R_{\text{thr}}$ as selected experts: $\mathcal{E} = \{E_\omega\}_\omega$. **return** $\mathcal{E}$
14: **end procedure**

As a variation of the EoE framework, the combination of experts using the endorsement concept but without any selection process is also introduced. This is called Endorsement-weighted EoE (EwEoE) framework. The output is calculated as:

$$D_{\text{EwEoE}} = \frac{1}{\sum_{\omega=1}^{N_e} r_\omega} \sum_{\omega=1}^{N_e} r_\omega D_\omega. \quad (5)$$

Also, for the purpose of comparison, traditional direct combination of experts is also considered. We denote the output of this method as Average EoE (AvgEoE):

$$D_{\text{AvgEoE}} = \frac{1}{N_e} \sum_{\omega=1}^{N_e} D_\omega. \quad (6)$$

## III. EXPERIMENTAL RESULTS

### A. Use case 1: Application to participants in the H-DIBCO'12

As the first example, the EoE framework was applied to the participants in the H-DIBCO'12 contest [11]. There were 23 participating methods in the H-DIBCO12 contest. Therefore, there are 23 experts in our ensemble. There was no need to have access to their code, as the outputs of all methods on the dataset images are published by the organizers and are available on the Internet.[3] All three EoE approaches to combine participants outputs were considered, and the summary of the results is provided in Table I.[4] It can be seen that the EoE performed more than 3% better than the $1^{st}$ rank in the contest in terms of F-measure (and 32% in terms of DRD). Also, as can be seen from the table, both proposed EoE and EwEoE frameworks achieved higher F-measure score compared to the AvgEoE approach. Actually, the performance of EwEoE is slightly better than that of EoE. Although the EwEoE framework gave a better combination of the experts outputs, it should be noted that the majority of experts (methods) participated in the H-DIBCO'12 have reached a common level of maturity, and therefore the ensemble is more uniform and stable. Therefore, expert selection among them would have less benefit. This can be seen from Figure 7 in which only one expert has been removed after consolidation. The actual outputs of three combing approaches are also shown in Figure 6. This might not be the case for other ensembles in which a large number of inefficient experts might have participated. In those types of ensembles, even a small weight assigned to that large number of irrelevant experts may shift and bias the final result and reduce the performance of the EwEoE framework. An example of that situation in provided in Section III-B. Also, it is worth noting that selection of the H-DIBCO'12 was because it was the most recent published dataset, and therefore there is no limitation in terms of handwritten or printed document images for the EoE framework. The performance of this framework on other datasets will be reported in future.

### B. Use case 2: Automated grid-based Sauvola method applied to the H-DIBCO'12 dataset

In practice, access to all experts who participated in the H-DIBCO'12 is not possible. Therefore, we decided to apply the EoE framework on a standard binarization method. For this purpose, the grid-based (Gb) Sauvola method [3] in its barehanded configuration without any preprocessing or postprocessing step was selected (see Figure 1(b)). The Gb Sauvola method has three parameters to set: $k$, $R$, and $s$. $k$ and $R$ are real numbers in $[0, 1]$ interval, and $s = 2 * G_s + 1$ is the corresponding scale where $G_s$ is the grid scale (for more information, see [3]). To build the ensemble of the experts, various combinations of these three parameters were considered. To be specific, those pair values of $k$ and $R$ that gave optimal binarization for the H-DIBCO'10 dataset were selected (see Table II). For $G_s$, the following values were used: $\{6, 9, 12, 15, 18, 24, 30\}$. The pairs in Table II and the set of $s$ values were then combined to generate an ensemble of 84 experts.

---
[3] http://utopia.duth.gr/~ipratika/HDIBCO2012/hdibco2012results.htm
[4] p-FM, PSNR, DRD, and MPM stand for pseudo F-Measure, Peak Signal-to-Noise Ratio, Distance Reciprocal Distortion Metric, and Misclassification Penalty Metric respectively: http://www.synchromedia.ca/web/reza/expres/obj_eval_code. Also, we introduce $FM_1$ as the average F-measure excluding the worse case. In addition, +% stands for percentage of difference with repsect to performance, i.e. it has an implicit sign-inversion for those metrics that decrease with increase in the performance, such as DRD and MPM.

TABLE I
THE PERFORMANCE OF VARIOUS COMBINING APPROACHES ON THE
H-DIBCO'12 CONTEST. THE FIRST RANK AND THE METHOD WITH
HIGHEST F-MEASURE ARE ALSO INCLUDED. $FM_1$ IS THE SAME AS
F-MEASURE EXCEPT IT IGNORES THE WORSE-CASE IMAGE.

| Method | F-measure | $FM_1$ | p-FM | PSNR | DRD | MPM |
|---|---|---|---|---|---|---|
| EoE H-DIBCO'12 | 92.53 | 93.09 | 95.06 | 20.39 | 2.32 | **0.30** |
| EwEoE H-DIBCO'12 | **92.76** | **93.31** | **95.15** | **20.52** | **2.26** | 0.37 |
| AvgEoE H-DIBCO'12 | 92.45 | 93.07 | 95.04 | 20.39 | 2.35 | 0.31 |
| $1^{st}$ rank [7] | 89.47 | — | 90.18 | **21.80** | 3.44 | 0.46 |
| Highest FM [8] | **92.85** | — | 93.34 | 20.57 | 2.66 | 0.72 |
| +% (EoE - $1^{st}$ rank) | 3.42% | — | 5.41% | −6.47% | 32.56% | 34.78% |
| +% (EoE - Highest FM) | −0.34% | — | 1.84% | −0.88% | 12.78% | 58.33% |

TABLE II
OPTIMAL $k$ AND $R$ VALUES FOR THE H-DIBCO'10 DATASET.

| | Optimal $k$ | Optimal $R$ | | Optimal $k$ | Optimal $R$ |
|---|---|---|---|---|---|
| 1 | 0.1 | 0.25 | 7 | 0.2444 | 0.4267 |
| 2 | 0.15 | 0.15 | 8 | 0.3389 | 0.25 |
| 3 | 0.15 | 0.25 | 9 | 0.4333 | 0.3056 |
| 4 | 0.15 | 0.3611 | 10 | 0.5278 | 0.3056 |
| 5 | 0.15 | 0.4167 | 11 | 0.6222 | 0.4167 |
| 6 | 0.15 | 0.75 | 12 | 0.8111 | 0.3611 |

TABLE III
THE EoE FRAMEWORK WITH THE Gb SAUVOLA METHOD ON THE
H-DIBCO'12 CONTEST.

| Method | FM | $FM_1$ | p-FM | PSNR | DRD | MPM |
|---|---|---|---|---|---|---|
| EoE Gb Sauvola | **85.95** | **86.59** | **91.61** | **17.83** | **4.64** | **0.79** |
| AvgEoE Gb Sauvola | 79.56 | 81.26 | 85.02 | 16.61 | 6.97 | 2.40 |
| EwEoE Gb Sauvola | 81.37 | 82.51 | 86.88 | 16.81 | 6.62 | 2.46 |
| Reported Sauvola [11] | 82.89 | — | 87.95 | 16.71 | 6.59 | — |
| +% (EoE - Reported) Sauvola | 3.69% | — | 4.16% | 6.70% | 29.59% | — |

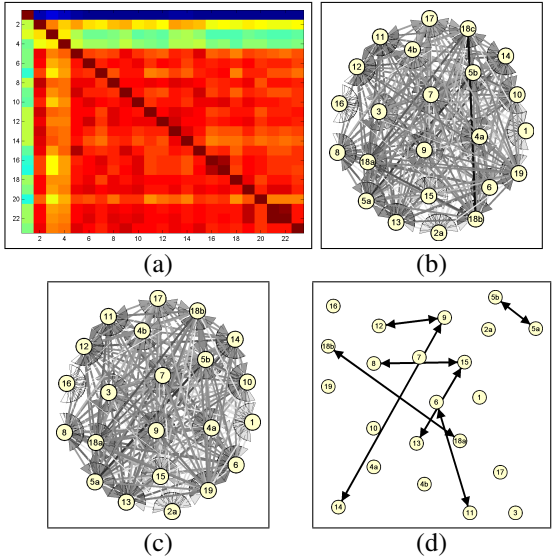

Fig. 5. An example of endorsement graph for the H-DIBCO'12 use case. a) The endorsement graph $R$ of image H12. b) The graph visualization of (a). c) The consolidated graph. d) The final selected experts and schools.

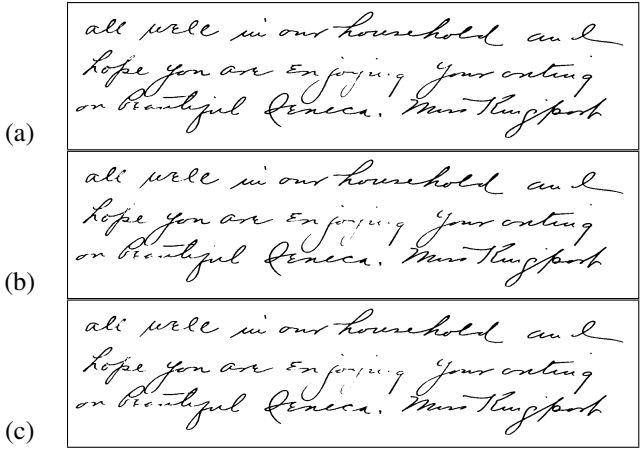

Fig. 6. a) the output of the EoE framework of image H12 in the H-DIBCO'12 contest use case. b) The same as (a) except using the EwEoE framework. c) The same as (a) except using the AvgEoE approach.

The ensemble of the Gb Sauvola experts were applied to the H-DIBCO'12 dataset, and the three approaches to combine the results were applied. The performance of the EoE framework and the other approaches is presented in Table III and compared with the reported performance of Sauvola method [11]. It can be seen that the EoE outperformed both the EwEoE and also the reported method in the literature. It is worth nothing that our proposed framework is highly objective. Even for selection of optimal $k$ and $R$ values, another dataset (H-DIBCO'10) was used. However, we want to emphasize that the participants in the contest did not have any access to the dataset at that time.

An example of consolidation and selection process of the EoE framework for image H12 and 84 Sauvola samples is shown in Figure 7. As can be seen, the number of experts has drastically reduced to 17 after consolidation. Also, 5 final schools of experts can be seen in Figure 7(d).

Figure 8 shows the shortfall of the EwEoE approach. As expected, the output of the EwEoE suffers from the bleed-through marks while the selection process of the EoE framework helps it to avoid a large number of irrelevant experts for this image.

*C. Use case 3: Automated Laplacian energy method method applied to the H-DIBCO'12 dataset*

In addition, the EoE framework was applied to the Laplacian energy method using the source code provided in [7]. In contrast to the use Case 1, here we are interested in several instances of the same method with different parameters, and therefore the source code was necessary. The ensemble of experts was the set of optimal parameter values for each individual image in the database. Because of lack of space, only the final results are reported in Table III. As can be seen, in terms of F-measure, the EoE version performs 1% higher than the best optimization algorithm, Alg. 3, provided in [7].

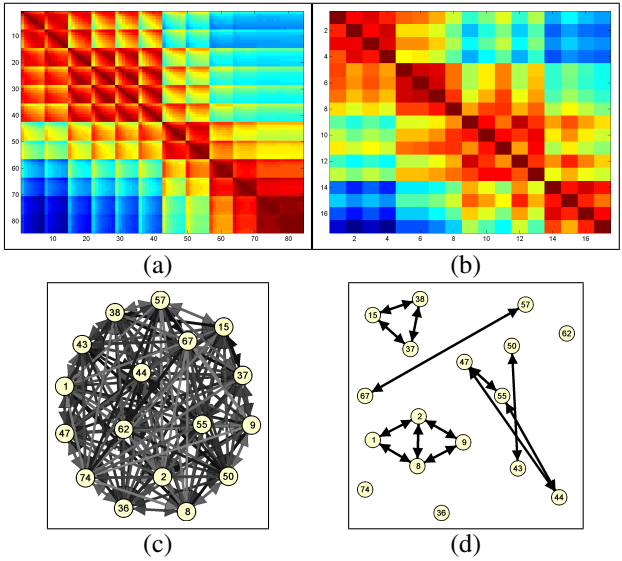

Fig. 7. An example of endorsement graph for the second use case. a) The endorsement graph $R$ of image H12. b) The consolidated graph. c) The graph visualization of (b). d) The final selected experts and schools.

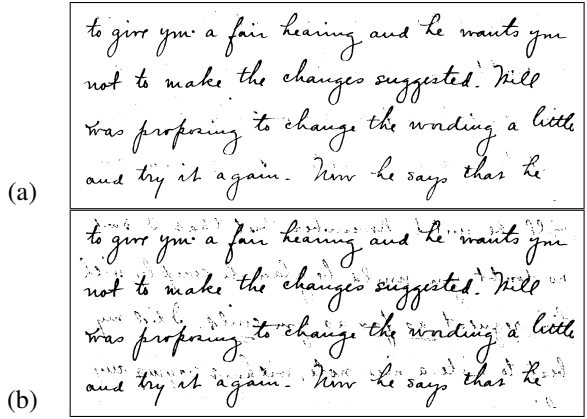

Fig. 8. a) The EoE output for the image H05 in the second use case. b) The EwEoE output of the same case.

TABLE IV
THE EoE FRAMEWORK WITH THE LAPLACIAN ENERGY METHOD ON THE H-DIBCO'12 CONTEST.

| Method | FM | $FM_1$ | p-FM | PSNR | DRD | MPM |
|---|---|---|---|---|---|---|
| EoE Laplacian energy | **94.78** | **95.23** | **95.60** | **21.97** | **1.81** | 0.72 |
| Laplacian (Alg. 3 in [7]) | 93.73 | 94.94 | 94.24 | 21.85 | 2.10 | **0.29** |
| +% (EoE - Alg. 3 [7]) Laplacian | 1.12% | 0.31% | 1.44% | 0.55% | 13.81% | $-148.28\%$ |

Finally, to have an objective of the potential of the EoE framework, some modified versions of this framework have been participated in the DIBCO'13 contest.

## IV. CONCLUSION

The EoE framework has been introduced to robustly combine the outputs from an ensemble of related and unrelated experts using consolidation and selection concepts. First, an endorsement graph is defined based on the confidentness of the experts. Then, following two generic principles, consolidation of saturated opinions and selection of schools of experts are performed. The opinions from the experts belonging to the final selected schools will be used to generate a robust combination. For the case of binarization methods, a confidentness map is defined using the local values of the stroke width. The framework was successfully applied to three use cases on the H-DIBCO'12 dataset. Although it was tested on a dataset of handwritten manuscripts because of limited space, the framework is general and can also handle printed documents.

Many aspects of the proposed framework (and its variations, such as the EwEoE framework) could be improved. For example, the process of consolidation (in particular, the selection of the consolidation threshold), definition of endorsement, and school identification process (for example, other clustering approaches other than thresholding) will be further investigated. The proposed EoE framework can be also used in other decision making problems. Two examples of those decision making environments are parliament setting and opinion fraud detection. However, caution should be taken when working with smart experts, such as humans. In those cases, some of them could use their awareness of the selection process to collectively adjust their behavior in order to make their associated alliance dominates the final result.